\documentclass[conference]{IEEEtran}
\IEEEoverridecommandlockouts
\usepackage{cite}
\usepackage{amsmath,amssymb,amsfonts}

\usepackage[utf8]{inputenc} 
\usepackage[T1]{fontenc}    
\usepackage{hyperref}       
\usepackage{url}            
\usepackage{booktabs}       
\usepackage{amsfonts}       
\usepackage{amsthm}
\usepackage{nicefrac}       
\usepackage{microtype}      
\usepackage{xcolor}         
\usepackage{algorithm} 
\usepackage{algpseudocode} 
\usepackage{graphicx}
\usepackage{derivative}
\usepackage{amsmath}

\begin{document}
\title{STORM: Strategic Orchestration of Modalities for Rare Event Classification}
\author{\IEEEauthorblockN{Payal Kamboj, Ayan Banerjee and Sandeep K.S. Gupta}
\IEEEauthorblockA{\textit{Impact Lab, Arizona State University} \\
Tempe, Arizona, USA \\
\{pkamboj, abanerj3, sandeep.gupta\}@asu.edu}

}

\maketitle

\begin{abstract}
In domains such as biomedical, expert insights are crucial for selecting the most informative modalities for artificial intelligence (AI) methodologies. However, using all available modalities poses challenges, particularly in determining the impact of each modality on performance and optimizing their combinations for accurate classification. Traditional approaches resort to manual trial and error methods, lacking systematic frameworks for discerning the most relevant modalities. Moreover, although multi-modal learning enables the integration of information from diverse sources, utilizing all available modalities is often impractical and unnecessary. To address this, we introduce an entropy-based algorithm STORM to solve the modality selection problem for rare event. This algorithm systematically evaluates the information content of individual modalities and their combinations, identifying the most discriminative features essential for rare class classification tasks. Through seizure onset zone detection case study, we demonstrate the efficacy of our algorithm in enhancing classification performance. 
By selecting useful subset of modalities, our approach paves the way for more efficient AI-driven biomedical analyses, thereby advancing disease diagnosis in clinical settings.

\end{abstract}

\begin{IEEEkeywords}
Deep Learning, Biomedical Imaging, Multimodality, Expert Knowledge, Modality Selection.
\end{IEEEkeywords}

\vspace{-0.1 in}\section{Introduction}
\IEEEPARstart{I}{n} the realm of Artificial Intelligence (AI), multimodal learning stands out as a powerhouse~\cite{NEURIPS2021_5aa3405a,kamboj2023transfer}. This approach doesn't just rely on one type of data but rather combines insights from various sources such as images, texts, signals etc~\cite{DBLP}. By doing so, multimodal learning enhances the overall adaptability and resilience of the model it employs. Literature review demonstrates that compared to a single-modal learning, not only multimodal learning consistently outperforms in real-world scenarios but also lowers the overall associated risks, making the learning process more stable~\cite{JMLRv25,bapna2022mslam,NEURIPS2021_5aa3405a,banerjee2024frameworkdevelopingevaluatingethical}. For instance, in autonomous vehicles (AVs), multimodal learning integrates data from diverse sensors such as Light Detection and Ranging (LiDAR), cameras, radar, GPS, and inertial measurement units (IMUs)~\cite{yeong2021sensor}. By fusing insights from these different modalities, AVs can make informed decisions in real-time, ensuring safer navigation through complex environments. However, as we venture deeper into the large-scale multimodal deep learning (DL), we face a significant challenge of efficient learning with such diverse data. The temptation might be to utilize all available modalities, but multimodal data, with its density and high dimensionality, can quickly escalate the complexity of models and may introduce redundant information ultimately confusing the overall AI method~\cite{JMLRv25}. Additionally, the more modalities we incorporate, the less incremental benefit we might derive, leading to diminishing returns~\cite{JMLRv25}. Hence, it becomes crucial to judiciously select the most relevant modalities, not only for computational efficiency but also to minimize the burden of maintaining unnecessary data streams. 

\begin{figure}
\centering
\includegraphics[width=1.1\columnwidth, clip=true,trim=0 0 0 0]{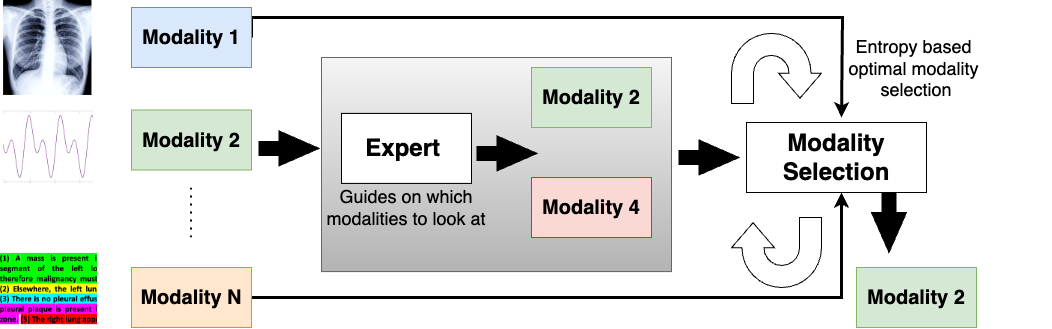}
\caption{Modality Selection: Within this framework, specific modalities, such as Modality 2 and Modality 4 as illustrated, are initially curated by domain experts. To assess the discriminative potential of each modality, an entropy-based technique is employed for selection. Furthermore, the remaining modalities can also undergo this evaluation to determine their respective contributions.}
\label{fig1as}
\end{figure}

The current digital health revolution seeks to enhance healthcare by using data from various sources~\cite{banerjee2024frameworkdevelopingevaluatingethical}~\cite{10.1145/3616388.3617550}~\cite{banerjee2024cpsllmlargelanguagemodel}. For example, the utilization of electronic medical records, radiology images, and genetic repositories to advance Cardiovascular Disease Care~\cite{amal2022use}. Similarly, in the domain of Seizure onset zone (SOZ) detection, fMRI and activation time series signals play pivotal roles~\cite{Kamboj2024,Banerjee2023},~\cite{kamboj2023mergingdeeplearningexpert}. Yet, a critical question is should we indiscriminately input all modalities into machine learning models, or should we selectively choose those that offer maximal information and discriminative features, avoiding the inclusion of modalities that contribute little to no performance improvement~\cite{kline2022multimodal}? Effectively addressing this query necessitates a method to quantitatively assess the potential contributions of each modality beforehand.

\subsection{Expert Knowledge based Modality Selection}
In biomedical domain, the initial selection of modalities often falls within the purview of domain experts. Take coronary artery disease (CAD) detection, for instance~\cite{mahmoodzadeh2011diagnostic}. While a plethora of modalities such as 12-lead Exercise Stress electrocardiogram (ECG) time-series signals and ECG images are available, experts can discern specific `N' leads to look at which are crucial for CAD detection. This expertise raises pertinent questions for AI: Are all `N' time-series signals selected equally important? Could the corresponding images alone, or in conjunction with the expert-selected leads, yield superior results? Similarly, in the context of SOZ detection, experts may advocate for examining both fMRI images and brain activation time series signals. However, do these modalities collectively furnish discerning insights for AI models? The crux lies in recognizing the pivotal role of expert knowledge in winnowing down the array of modalities, albeit acknowledging its inherent fuzziness and potential to amplify intra-class variance~\cite{KambojTAI,hossain2023edgcon}. Moreover, what about modalities not endorsed by experts? Do they furnish redundant, incremental, or genuinely novel insights? Addressing these queries necessitates thorough evaluation using a unified framework.

\subsection{Overview of Solution}
~\textit{``Rare events are extremely infrequent events whose characteristics make them highly valuable.
Such events appear with extreme scarcity and are hard to
predict, although they are expected eventually''}~\cite{sokolova2010evaluation}. Despite the rarity, these instances are highly significant as they contain crucial information. Entropy represents the level of uncertainty within the data. Greater randomness correlates with higher entropy. Information gain leverages entropy to guide decision-making. When entropy decreases, information content increases. In literature, class-wise entropy has been used to quantify class imbalance~\cite{8703114}. However, this method not only considers the imbalance in the number of instances for each class but also focuses on the relative importance of a sample in the information content of the dataset which can help in modality selection. To this end, we explore the intricacies of selecting modalities in multimodal learning, aiming to identify the optimal combination that maximizes learning performance from a given set of input modalities. This process involves two key steps: first, expert advice is used to narrow down the modalities and select relevant features within each modality; second, an entropy-based algorithm is applied to select the optimal modality from the expert's suggestions and remaining modalities (if any)for rare event detection(Fig.~\ref{fig1as}). Two major challenges are: the lack of direct methods to evaluate the learning effectiveness of a modality set, and the NP-hard nature of selecting the optimal subset. To address these, we propose a unified theoretical framework to quantitatively evaluate the learning efficacy of different modality sets.

\subsection{Application Domain}

In this study, we focus on a critical medical application: the detection of SOZ using resting-state fMRI. This is vital for identifying the brain region responsible for seizures in individuals with pharmaco-resistant focal epilepsy (PFE)~\cite{Banerjee2023, Kamboj2024}. Independent Component Analysis (ICA) performed on rs-fMRI divides the data into two modalities: spatial (images) modality and temporal (time-series) modality. ICA typically produces between 100 to 200 Independent Components (ICs) in both the modalities, with only a small fraction (less than 5\%) being SOZ ICs, rare class~\cite{Kamboj2024}. This also makes SOZ detection problem an imbalanced class problem. Despite ICs being orthogonal, ICA cannot categorize them as Resting State Networks (normal brain function), SOZ, or noise (measurement artifacts). Consequently, expert manual sorting of these ICs data in three classes RSN, noise and SOZ is necessary, but it is time-consuming, subjective, and hampers the reproducibility and accessibility of rs-fMRI-based SOZ identification, and hence the requirement of automated SOZ detection technique. As noted in ~\cite{Banerjee2023, Kamboj2024}, leveraging expert knowledge has proven to enhance SOZ classification, with specific rules proposed for both modalities. Here, we denote the raw spatial modality as Basic Modality, expert knowledge suggested modalities as derived modalities. Derived modalities are expert-suggested ways to transform or reinterpret the raw spatial or temporal data, such as clustering in the spatial domain or frequency analysis in the temporal domain, which can then yield specific features like the number of clusters or their overlap with brain regions. Derived spatial modality is further categorized as  Modality D1 and the derived temporal modality as Modality D2. The expert rules within derived modalities are outlined as follows:\\
\vspace{-0.1mm}
\noindent{\bf Modality D1a:} SOZ spatial component ideally has one cluster~\cite{Banerjee2023, Kamboj2024}.

\noindent{\bf Modality D1b:} SOZ has activation extended from grey matter to the ventricles through the white matter~\cite{Banerjee2023, Kamboj2024}. 

\noindent{\bf Modality D2a:} SOZ signal power spectra exhibit dominant frequencies greater than 0.07 Hz, and spikes have sparse, transient representation in activelet basis~\cite{Banerjee2023, Kamboj2024}. 

\noindent{\bf Modality D2b:} The rs-fMRI SOZ expected to demonstrate sparse representation in sine dictionary at frequencies higher than those found in RSN (0.01-0.1Hz), and the BOLD time series may display irregular patterns~\cite{Banerjee2023, Kamboj2024}.

Note than even though derived modality D1 (spatial) and derived modality D2 (temporal) are two different modalities, the derived rules within each of these modalities denoted as D1a, D1b and D2a, D2b are rules based features extracted from modality D1 and modality D2 respectively. Based on the basic modality, and expert knowledge based derived modalities D1 and D2, our framework aims to determine whether each modality individually contributes discriminative and significant information, or if one modality is crucial while the other merely offers marginal performance improvements that may not justify the computational resources required. \\
\vspace{-1.00mm}
\section{Methodology}

The methodology comprises two main components:\\
i) Assessment: Evaluating existing class imbalances and the relative importance of sample information within each modality.\\
ii) Selection: Modalities are selected that collectively minimize entropy imbalance, ensuring they provide the most discriminative and representative features, effectively reducing class imbalance while preserving critical information.

 \subsection{ Modality Assessment}
 Class-wise entropy, defined in \cite{8703114}, is a method which not only considers the imbalance in the number of instances for each class but also focuses on the relative importance of a sample in the information content of the dataset. This method requires a distance definition $dist(x_i, x_j)$, between representations $x_i$ and $x_j$ of two instances of raw data from a modality $d$ $y^d_i$ and $y^d_j$ in the dataset $Y^d$ with $n$ instances. The representations are function of a classifier $M_d$, which is trained on data from a modality $d$. 
 
 A modality $d$ can be either: i) a basic modality, in other words raw data from some modality such as image or signals, or ii) a derived modality, which is expert-suggested way to transform the raw data in a modality $d$.
 
 Each instance in a given modality can belong to a unique class out of a finite number ($m$) of classes $c_r \in \{c_1, \ldots, c_m\}$. For each instance $y_i$, a set $Q(x_i)$ is derived, which is the set of all instances $y_j$ such that $x_j, x_i \in c_r$, and $y_j$ is a member of the $K$ nearest neighbor set of $y_i$ using the representations $x_i$ and $x_j$, and distance metric $dist(x_i,x_j)$. The set $Q(x_i)$ measures the density of $y_i$, $\lambda(x_i)$ using Eqn.~\ref{eqn:dist}.

\begin{equation}
\scriptsize
\label{eqn:dist}
\lambda(x_i) = \frac{1}{|Q(x_i)|} \sum_{j=1}^{|Q(x_i)|} \frac{1}{\text{dist}(x_i, x_j)},
\end{equation}
where \(|Q(x_i)|\) is the number of elements in the set \(Q(x_i)\). For each instance \(y_i\) of a class \(c_r\), the class average density is computed using Eqn. \ref{eqn:dist2}.

\begin{equation}
\scriptsize
\label{eqn:dist2}
\gamma(x_i) = \frac{\lambda(x_i)}{\sum_{j=1}^{|c_r|} \lambda(x_j)},
\end{equation}
where \(|c_r|\) is the number of elements of class \(c_r\). The class entropy for a class \(c_r\) is then defined using Eqn. \ref{eqn:theta},
\begin{equation}
\scriptsize
\label{eqn:theta}
\theta_r = \sum_{i=1}^{|c_r|} (-\gamma(x_i) \log_2 \gamma(x_i)).
\end{equation}

A significant discrepancy in class entropy from Eqn. \ref{eqn:theta} indicates class imbalance. For two classes $c_r$ and $c_s$, if $\theta_r > \theta_s$, then a single instance from \(c_r\) has more information content than that from \(c_s\). This implies that the representation \(x_i\) for an instance from class \(c_r\) is not representative of the class. Consequently, the removal of the sample will result in the loss of information that cannot be learned using the representation \(x_j\) of other instances in class \(c_r\). Therefore, either \(c_r\) needs more samples or needs a different representation. 
We investigate class imbalance in raw data's basic modality, 
and expert knowledge based representation using both derived Modality D1 and D2. The distance metric $dist(.,.)$ for all cases is Euclidean distance. For raw data, we used the peak signal to noise ratio (PSNR) as the representation for each instance. For raw data, we also tried DL's penultimate layer representation of VGG 16 deep CNN model \cite{b60} as the representation $x_i$ for a class $c \in \{RSN, NOISE, SOZ\}$. For expert knowledge, we used the features shown as Modality D1a, Modality D1b, Modality D2a and Modality D2b as representation $x_i$.  
It was observed that both raw data and CNN intermediate representation of raw data have significant discrepancy in class entropy across classes. However, with expert knowledge, the discrepancy between class entropy is significantly reduced. This suggests that the expert knowledge based modalities extracted represent exemplary SOZ characteristics and helps alleviate class imbalance problem.

\vspace{-1mm}
 \subsection{Modality Selection}


Formally, for a classification problem with $m$ original classes $C = \{c_1 \ldots c_m\}$, there can be a set of trained classifiers $\mathcal{M}$ on different modalities. In this paper, we use classifier and machine terminology interchangeably. Each classifier $M_d \in \mathcal{M}$, takes the raw data $Y$ as input and divides into partitions with the label set $S^{M_d} \subset 2^C$, such that each label $s^{M_d}_k \in S^{M_d}$ meets the following criteria:

\begin{scriptsize}
\begin{eqnarray}
&&\forall  k,l \in \{1 \ldots |S^{M_d}|\}, k\neq l, s^{M_d}_k \bigcap s^{M_d}_l = \phi \text{ mutually exclusive }\\\nonumber
&&\bigcup^{|S^{M_d}|}_{k = 1}{s^{M_d}_k} = C \text{ exhaustive }\\\nonumber
&&\forall  s^{M_d}_k \in S^{M_d} \exists Z \subset C : s^{M_d}_k = \bigcup_{z \in Z} z \text{ formed with union of original labels } 
\end{eqnarray}
\end{scriptsize}
Here $\phi$ is the null set. Each classifier $M_d$ has several intermediate representations $x_i$ of the raw data $y_i \in Y$, we consider the ``most discriminativ'' representation $\mathcal{F}_{M_d}: Y \rightarrow \mathcal{R}^b$, where $b$ is the representation dimension. There can be several definitions of most discriminative representation, including the difference between intra-class and inter-class distance using the distance function $dist(.,.)$ in Eqn. \ref{eqn:dist}. 

The discriminative feature function $\mathcal{F}_{M_d}$ can be used to represent each raw data in the original class set $C$, regardless of the partitions used in $M_d$ during training. We utilize this representation in Eqn. \ref{eqn:dist} to compute a new $\lambda^{M_d}(x_i)$ by replacing each instance $x_i$ by $\mathcal{F}_{M_d}(y_i)$. Following the entropy calculation in Eqn. \ref{eqn:theta} we can derive the entropy $\theta^{M_d}_r$ for each classifier $M_d$ and for each original class $c_r \in C$. We define the \textit{entropy imbalance metric} as:

\begin{equation}
\scriptsize
\label{eqn:entropyImbalance}
    \eta^{M_d} = \max_{\forall c_r \in C}{ \theta^{M_d}_r - E(\theta^{M_d}_r)} 
\end{equation}

Ideally the best classifier should have an intermediate representation that has the lowest value of $\eta^{M_d}$, since it implies that representative class features were learned. This metric can then be used in a Hunt's algorithm~\cite{Hunt}, to develop a decision tree that dictates the modality selection strategy.

This strategy evaluates the \textit{entropy imbalance gain} $EIG(M_d)$ achieved by a classifier $M_d$ using Eqn. \ref{eqn:gain}.
\begin{equation}
\scriptsize
\label{eqn:gain}
EIG(M_d) = \eta^R - \eta^{M_d},
\end{equation}
where $\eta^R$ is the entropy imbalance of the raw data. \\

\noindent{\bf STORM} (\textbf{St}rategic \textbf{Or}chestration of \textbf{M}odalities for Rare Event classification)  algorithm overview: To select discriminative modalities for a SOZ rare class $c_r$, we use the Algorithm \ref{alg:eksaii}: It takes three configuration parameters: a) entropy imbalance threshold $\epsilon_m$, used to determine that classifiers $M_1$ and $M_2$ are equivalent if $abs(EIG(M_1)-EIG(M_2)) < \epsilon_m$, b) impurity threshold $\epsilon_g$, used to determine if a classifier results in poor classification using Gini Index~\cite{sitthiyot2020simple} (gini index of the classifier output is greater than $\epsilon_g$ ), and c) dependability threshold $d_{th}$, that is used to set a preference to a given classifier. Algorithm \ref{alg:eksaii} also takes the training data and a set of classifiers $\mathcal{M}$ as input. It then runs the following steps: 

\noindent{\bf Step 1:} It chooses a classifier with the maximum EIG. 

\noindent{\bf Step 2:} If the class set $S^{M_d}$ of the classifier contains the rare class $c_r$, then it evaluates intra-class variability through Gini Index.

\noindent{\bf Step 3:} If Gini index $<\epsilon_g$ then the algorithm stops. Else it repeats Step 1 with instances only from class $s^{M_d}_s=c_r$ to find another classifier that can be cascaded with the $M_d$. 

\noindent{\bf Step 4:} If no label matches $c_r$, then the algorithm searches for a label set $s^{M_d}_j$ such that $c_r \subset s^{M_d}_j$, sets the training samples to the samples from the class labeled $s^{M_d}_j$ and restarts from Step 1. If there is a tie between classifiers, then the classifier with confidence score $> d_{th}$ is used to compute $EIG(M_d)$. 

\noindent{\bf Stopping condition:} The process continues until the training set is exhausted or validation accuracy remains unchanged for consecutive cycles.

\begin{algorithm}
\caption{STORM Algorithm}
\label{alg:eksaii}
\scriptsize
\textbf{Input:} Raw data $Y$, Rare class $c_r$, Thresholds $\epsilon_{m}$, $\epsilon_g$, Dependability threshold $d_{th}$, set of classifiers $\mathcal{M}$ such that the modified class labels for classifier $M_d$ is $S^{M_d}$.

\begin{enumerate}
    \item Sample set $\Psi = Y$
    \item \textbf{While} $\Psi$ is not empty and there is a significant change in validation accuracy:
    \begin{enumerate}
        \item For each classifier $M_d \in \mathcal{M}$:
            \begin{itemize}
                \item Compute $EIG(M_d)$ from Eqn. \ref{eqn:gain} on the set $\Psi$
            \end{itemize}
        \item Choose classifier with maximum gain: $M_d \leftarrow \arg\max_{M_d} EIG(M_d)$
        \item \textbf{If} there is no tie in $EIG(M_d)$ within threshold $\epsilon_{m}$:
            \begin{itemize}
                \item \textbf{If} $\exists s^{M_d}_s$ in $S^{M_d}$ such that $s^{M_d}_s = c_r$:
                    \begin{itemize}
                        \item Compute purity of partition $s^{M_d}_s$ using Gini index
                        \item \textbf{If} Gini index $>$ $\epsilon_g$:
                            \begin{itemize}
                                \item Restart from Step 2 with $\Psi = s^{M_d}_s$
                            \end{itemize}
                        \item \textbf{Else}: Stop
                    \end{itemize}
                \item \textbf{Else}:
                    \begin{itemize}
                        \item GOTO Step 2 with instances from partition $\Psi = s^{M_d}_j \in S^{M_d}$ such that the original class label $c_r \subseteq s^{M_d}_j$
                    \end{itemize}
            \end{itemize}
        \item \textbf{ElseIf} there is a tie between $M_1$ and $M_2$:
            \begin{itemize}
                \item Compute confidence scores for classifiers $M_1$ and $M_2$
                \item Choose classifier with score $> d_{th}$
                \item Repeat Steps 8 through 16
            \end{itemize}
    \end{enumerate}
\end{enumerate}
\end{algorithm}

\begin{figure}
\centering
\includegraphics[width=0.75\columnwidth, clip=true,trim=0 0 0 0]{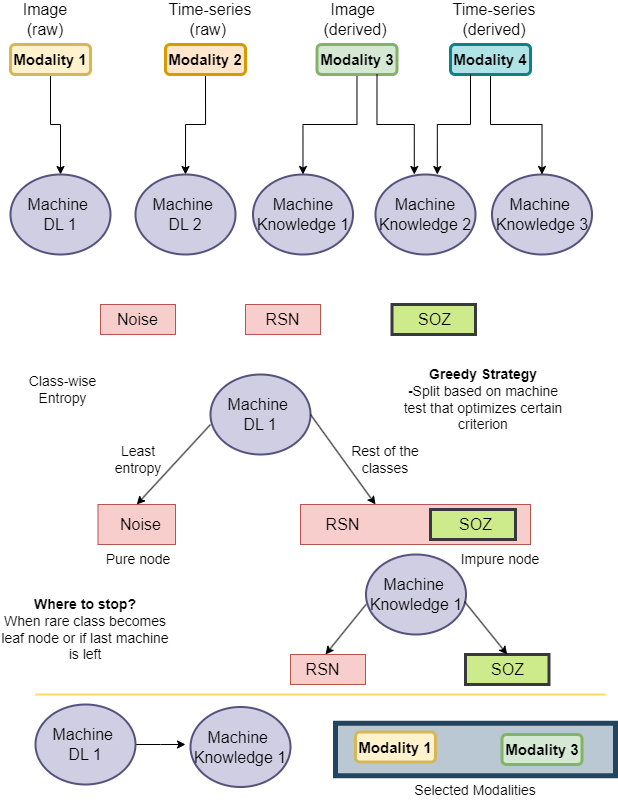}
\caption{SOZ classification using STORM.}
\label{fig2}
\end{figure}
Utilizing algorithm~\ref{alg:eksaii}, we identify that among the provided basic and derived modalities, one basic modality and one derived modality D1 was deemed informative, while the remaining derived D2 modality was deemed unnecessary (Figure~\ref{fig2}). We employed 2D-CNN for the basic image modality's classifier and Support Vector Machine (SVM) for the two derived modalities D1 and D2 which had their own respective features. The CNN was tasked with classifying noise and non-noise ICs, while the SVM was employed to classify SOZ and RSN ICs. To address class imbalance, Synthetic Minority Over-sampling Technique (SMOTE) was applied before SVM~\cite{SMOTE}. Subsequently, the final classification labels were determined by considering outputs from both the CNN and SVM models. Specifically, CNN's noise labels replaced all SVM labels except in cases where the SVM's classification score exceeded 0.9 threshold score for SOZ classification. For non-noise labels from the CNN, SVM labels were adopted as the final classification labels.

\section{Data and Results}

 \subsection{Data collection} The retrospective analysis for this project received approval from the IRB (IRB 20-358) of Phoenix Children's Hospital (PCH). We obtained a rs-fMRI dataset comprising 52 children,aged between 3 months and 18 years, diagnosed with epilepsy. 
 Scans were conducted using a 3T MRI unit, with further technical specifications available in ~\cite{Banerjee2023, Kamboj2024}

 \subsection{Experiments and Results}
To assess the impact of each modality on the performance of identifying the SOZ, we conducted experiments incorporating various configurations. These experiments encompassed scenarios where all basic and derived modalities were utilized, as well as cases where only Modality D1, Modality D2, basic modality, or derived modality were exclusively employed. Furthermore, for a comprehensive analysis, we examined the performance within expert-suggested derived modalities by excluding the expert-suggested rule in a feature form within specific modalities one by one. 
The configurations for experiments are broadly categorized as follows:

\noindent{\bf Comprehensive SOZ Detection:} This evaluates the model's performance on all available modalities.

\noindent{\bf Basic Modality Exclusion:} We examine the model's performance with only basic modality.

\noindent{\bf Expert-Suggested Modality Exclusion:} We assess the model's performance by incorporating modalities suggested by experts.

\noindent{\bf Modality D1 Exclusion:} Expert suggested spatial modality is omitted, resulting in three distinct configurations: a) omission of D1 Modality's feature D1a, b) omission of D1 Modality's feature D1b, and c) exclusion of the entire Modality D1.\\
\noindent{\bf Modality D2 Exclusion:} Expert derived temporal modality D2 is removed from the model. We explore three distinct configurations here: a) omission of D2 Modality's feature D2a, b) omission of D2 Modality's feature D2b, and c) exclusion of the entire Modality D2.
\vspace{-0.3mm}
\begin{table}[htbp]
\caption{Summary of SOZ Detection Results Across Various Modalities, Including modality feature ablation analysis, N=52} 
\scriptsize
\begin{center}
\begin{tabular}{|p{1.1 in}|p{0.3 in}|p{0.3 in}|p{0.3 in}|p{0.65 in}|}
\hline
\textbf{SOZ Detection with}& \textit{Accuracy} & \textit{Precision} & \textit{Sensitivity}  &\textit{F1 score} \\ 
\hline
All modalities included & 84.6\%        & 93.6\%        &89.7\%      &91.6\%    \\

\hline
Modality Basic excluded &  50.0\%         &89.6\%         &53.6\%     & 67.0\%    \\ 
    \hline
    
Modality Derived excluded & 46.1\%         &88.8\%         &48.9\%     & 63.0\%    \\ 
    \hline
 Modality D1  excluded & 0\%        & 0\%        & 0\%      &0\%    \\
   \hline
   Modality D1a excluded & 75.0\%        & 92.8\%        & 79.5\%     &85.6\%    \\
    \hline
     Modality D1b excluded & 0\%        & 0\%        & 0\%      &0\%    \\
    
    \hline
     Modality D2 excluded & 84.6\%         & 93.6\%         & 89.7\%    &91.6\%     \\ 
     \hline
    Modality D2a excluded &  84.6\%         & 93.6\%         & 89.7\%     &91.6 \%    \\ 
    \hline
     Modality D2b excluded &  84.6\%         & 93.6\%         & 89.7\%    &91.6 \%     \\ 
     \hline

  \end{tabular}
  \label{tbl:AblationStudy}
\end{center}
\vspace{-0.2 in}\end{table}

    
   


The results presented in Table ~\ref{tbl:AblationStudy} demonstrate the significance of different modalities in SOZ detection. Standard metrics used in~\cite{Banerjee2023,Kamboj2024} are used for SOZ detection. We found that removing the basic modality led to a notable decrease in the F1 score, dropping from 91.6\% to 67.0\%. Similarly, exclusion of derived modalities resulted in a reduction of the F1 score to 63\%, underscoring the importance of both basic and expert-derived modalities. Delving deeper into the evaluation of expert-derived modalities, it becomes evident that not all suggested modalities hold equal importance. For instance, removing spatial-derived modality D1 resulted in a significant drop in both accuracy and F1 score, emphasizing its crucial role in SOZ detection. Conversely, excluding the expert-derived temporal modality D2 yielded similar accuracy and F1 scores as it was when this modality was incorporated, indicating that while these temporal modalities might contribute to manual SOZ decision-making according to experts, they do not provide discriminative information from AI perspective. Hence, instead of blindly using all available modalities, only basic and derived D1 modalities are of utmost importance here.
\subsection{Another Case Study}
Another potential application of the proposed solution is CAD detection~\cite{b2}. Exercise stress electrocardiography (ECG) serves as a non-invasive, cost-effective tool for initial CAD assessments. In this context, the basic modality is ECG images, while an expert-derived modality is time-series signals extracted from these images. From the extracted time-series signals, valuable features can be identified, such as the Inferior (LII, LIII, aVF) and lateral (V5, V6) leads, which are critical for CAD detection. Using STORM, it can be assessed which modality is adequate.
\section{Conclusions}
This study highlights the importance of a systematic modality orchestration for rare event classification. We propose an entropy-based algorithm to identify the most informative modalities for accurate disease diagnosis. A case study on SOZ detection demonstrates the approach’s effectiveness, revealing that not all modalities, including expert-suggested ones, equally enhance performance.
This work sets a precedent for future research aimed at optimizing modality selection strategies, ultimately leading to improved healthcare outcomes through enhanced data-driven decision-making.
\section*{Acknowledgements}
We thank Varina Boerwinkle, Sarah Wyckoff and Bethany Sussman for introducing us to the case study and PCH data. The work is partly funded by DARPA AMP-N6600120C4020, DARPA FIRE-P000050426, and Helmsley Charitable Trust - 2-SRA-2017-503-M-B.

\bibliographystyle{ieeetr}
\bibliography{Asilomar}

\end{document}